\documentclass{article}
\pdfoutput=1

\PassOptionsToPackage{numbers, compress}{natbib}
\usepackage{amssymb,array,delarray,verbatim,alltt,epsfig,float,amsthm}
\usepackage{graphicx,algorithmicx,algorithm,algpseudocode,xspace,xcolor}
\usepackage{url}
\usepackage{epsf,epsfig,dsfont}
\usepackage{amssymb,amsmath}
\usepackage{pgf,amsbsy,amsfonts,amsmath,multicol,natbib}
\usepackage{graphicx,xspace,color,cancel}
\usepackage{tikz}
\usetikzlibrary{arrows,shapes,backgrounds,through,shadows,snakes}
\usetikzlibrary{decorations.pathmorphing}
\usetikzlibrary{calc}
\usetikzlibrary{matrix}

\graphicspath{{./images/}{./matlab/}{./figures/}{../matlab/}{../matlab/figures/}{../cup_first_edition/figures/}{./../codeOO/misc/}{../cup_first_edition_corrections/figures/}{./../codeOO/DemosExercises/}{../../projects/deepbelief/}
{../../talks/CambSciFest/figures/}{../../talks/CambSciFest/images/}}

\tikzstyle{bigdot}=[circle,draw=blue!80,fill=blue!25,anchor=center,align=center,minimum size=6mm]
\tikzstyle{bigemptydot}=[circle,draw=blue!80,anchor=center,align=center,minimum size=6mm]

\tikzstyle{biggreendot}=[circle,draw=green!80,fill=green!25,anchor=center,align=center,minimum size=6mm]
\tikzstyle{bigreddot}=[circle,draw=red!80,fill=red!25,anchor=center,align=center,minimum size=6mm]

\tikzstyle{dot}=[circle,draw=orange!80,fill=orange!25,anchor=center,align=center,minimum size=4mm]
\tikzstyle{emptydot}=[circle,draw=orange!80,anchor=center,align=center,minimum size=4mm]

\tikzstyle{reddot}=[circle,draw=red!80,fill=red!25,anchor=center,align=center,minimum size=4mm]

\tikzstyle{greendot}=[circle,draw=green!80,fill=green!25,anchor=center,align=center,minimum size=4mm]

\tikzstyle{bluedot}=[circle,draw=blue!80,fill=blue!25,anchor=center,align=center,minimum size=4mm]

\tikzstyle{neur}=[rectangle,draw=green!50,fill=green!50,minimum size=6mm,line width=2pt,>=stealth]  % continuous

\tikzstyle{fact}=[fill,minimum size=1.5mm,line width=2pt,>=stealth]

\tikzstyle{cont2}=[circle,draw=black!50,top color=white, % a shading that is white at the top...
bottom color=lilla!50!black!20,thick,minimum size=6mm,line width=1pt,>=stealth]  % continuous  node

\tikzstyle{contredb}=[circle,draw=red,top color=red, % a shading that is white at the top...
bottom color=red!50!black!20,thick,minimum size=8.1mm,line width=1pt,>=stealth]  % continuous  node
\tikzstyle{contyellowb}=[circle,draw=yellow,top color=yellow, % a shading that is white at the top...
bottom color=yellow!50!black!20,thick,minimum size=8.1mm,line width=1pt,>=stealth]  % continuous  node
\tikzstyle{contblueb}=[circle,draw=blue,top color=blue, % a shading that is white at the top...
bottom color=blue!50!black!20, thick,minimum size=8.1mm,line width=1pt,>=stealth]  % continuous  node

\tikzstyle{contred}=[circle,draw=red,top color=red, % a shading that is white at the top...
bottom color=white,thick,minimum size=6mm,line width=1pt,>=stealth]  % continuous  node
\tikzstyle{contyellow}=[circle,draw=yellow,top color=yellow, % a shading that is white at the top...
bottom color=yellow!50!black!20,thick,minimum size=6mm,line width=1pt,>=stealth]  % continuous  node
\tikzstyle{contblue}=[circle,draw=blue,top color=blue!80!black!50, % a shading that is white at the top...
bottom color=white, thick,minimum size=6mm,line width=1pt,>=stealth]  % continuous  node
\tikzstyle{contgreen}=[circle,draw=green,top color=green!80!black!50, % a shading that is white at the top...
bottom color=white, thick,minimum size=6mm,line width=1pt,>=stealth]  % continuous  node
\tikzstyle{contwhite}=[circle,draw=white,color=white, thick,minimum size=6mm,line width=1pt,>=stealth]  % continuous  node
\tikzstyle{contwhiteb}=[circle,draw=white,color=white, thick,minimum size=7.5mm,line width=1pt,>=stealth]  % continuous  node

\tikzstyle{cont}=[circle, draw,% a shading that is white at the top...
thick,minimum size=6mm,line width=1pt,>=stealth]  % continuous  node

\tikzstyle{contb}=[circle,draw=black!50,top color=white, % a shading that is white at the top...
bottom color=darkgreen!50!black!20,thick,inner sep=0pt,minimum size=8.1mm,line width=1pt,>=stealth]  % continuous  node

\tikzstyle{ocont}=[ellipse,draw=blue!50,thick,minimum size=6mm,>=stealth]  % continuous  node
\tikzstyle{blackcont}=[circle,draw=black!50,thick,minimum size=6mm,line width=2pt,>=stealth]  % continuous  node
\tikzstyle{oval}=[ellipse,draw=blue!50,thick,minimum size=6mm,line width=1pt,>=stealth]  % continuous node
\tikzstyle{ovalb}=[ellipse,draw=blue!50,thick,minimum size=7.5mm,line width=1pt,>=stealth]  % continuous node
\tikzstyle{disc}=[rectangle,draw=blue!50,thick,line width=1pt,minimum size=6mm]  % discrete node
\tikzstyle{obs}=[fill=blue!20,thick]  % observed node
\tikzstyle{opt}=[star,draw=red!50,thick,minimum size=6mm]  % decision node

\tikzstyle{fillred}=[fill=red!20,thick]  % observed node
\tikzstyle{fillgreen}=[fill=green!20,thick]  % observed node
\tikzstyle{purered}=[fill=red]  % observed node
\tikzstyle{state}=[rectangle,fill=red!20]  % state
\tikzstyle{sobs}=[fill=green!15,thick]  % sequentally observed node
\tikzstyle{fact}=[fill,minimum size=1.5mm,line width=2pt,>=stealth]
\tikzstyle{varfact}=[draw,minimum size=1.5mm,line width=2pt,>=stealth]
\tikzstyle{sep}=[rectangle,draw=magenta!50,thick,minimum size=6mm]  % discrete node
\tikzstyle{det}=[fill=red!15,rectangle,draw=red!50,thick,minimum size=6mm]  % deterministic node

\tikzstyle{dethid}=[diamond,draw=red!50,thick,minimum size=6mm]  % deterministic  hidden node

\tikzstyle{lineball}=[fill,-*,draw=red!50,line width=1.5pt]
\tikzstyle{redball}=[mark=*,mark options={fill=red!50,draw=red},mark size=0.5pt]
\tikzstyle{greenball}=[mark=*,mark options={fill=green!50,draw=green},mark size=0.5pt]
\tikzstyle{hid}=[circle,draw,thick]  %  non observed node

\tikzstyle{dec}=[rectangle,draw=red!50,thick,minimum size=6mm]  % decision node
\tikzstyle{utility}=[diamond,draw=red!50,thick,minimum size=6mm]  % utility node
\tikzstyle{contdec}=[circle,draw=blue!50,thick,fill=blue!10,line width=2pt]  % observed node after a decision
\tikzstyle{decutility}=[diamond,draw=red!50,thick,minimum size=6mm]  % utility node

\tikzstyle{contobs}+=[cont]
\tikzstyle{contobs}+=[obs]
\tikzstyle{discobs}+=[disc]
\tikzstyle{discobs}+=[obs]

\tikzstyle{obsred}+=[obs]
\tikzstyle{obsred}+=[red]

\tikzstyle{background grid}=[draw, black!50,step=.1cm]
\tikzstyle{dgraph}=[->, line width=1.5pt]
\tikzstyle{ugraph}=[line width=1.5pt]

\definecolor{magenta}{cmyk}{0.1,1,1,0.5}
\definecolor{darkgreen}{cmyk}{0.6,0.1,0.6,0.6}
\definecolor{pink}{cmyk}{0.1,1,1,0.1}
\definecolor{azzurro}{cmyk}{0.9333, 0.2471, 0.5569, 0.102}
\definecolor{lilla}{rgb}{0.5, 0, 0.5}
\definecolor{mygreen}{rgb}{0, 0.5, 0}
\definecolor{darkorange}{rgb}{1, 0.4, 0} % is probably not visible...
\definecolor{darkred}{rgb}{0.8, 0, 0}

\newcommand{\red}[1]{{\color{red}{#1}}}
\newcommand{\blue}[1]{{\color{blue}{#1}}}

\newcommand{\eframe}{\end{frame}}

\newcommand{\bmp}[1]{\begin{minipage}{#1}}
\newcommand{\bmpp}[2]{\begin{minipage}[#1]{#2}}
\newcommand{\emp}{\end{minipage}}
\newcommand{\cb}[1]{\left\{ {#1} \right\}}
\newcommand{\br}[1]{\left( {#1} \right)}

\newcommand{\sett}[1]{\myset{#1}}

\newcommand{\myset}[1]{\mathcal{\uppercase{#1}}}
\newcommand{\beq}{\[}
\newcommand{\eeq}{\]}

\tikzstyle{celim}=[circle,draw=red!25,thick,minimum size=6mm,line width=2pt,>=stealth]  %
\tikzstyle{delim}=[draw=red!25]  %

\newcommand{\btz}{\begin{tikzpicture}}
\newcommand{\etz}{\end{tikzpicture}}

\input{tikzlibrarybayesnet.code.tex}
\usepackage{wrapfig}

\newcommand{\wl}{\rho}

\usepackage[preprint]{neurips_2022}

\usepackage[utf8]{inputenc} % allow utf-8 input
\usepackage[T1]{fontenc}    % use 8-bit T1 fonts
\usepackage{hyperref}       % hyperlinks
\usepackage{url}            % simple URL typesetting
\usepackage{booktabs}       % professional-quality tables
\usepackage{amsfonts}       % blackboard math symbols
\usepackage{nicefrac}       % compact symbols for 1/2, etc.
\usepackage{microtype}      % microtypography
\usepackage{xcolor}         % colors
\usepackage{multirow}
\usepackage{float}
\usepackage{caption}
\usepackage{subcaption}
\usepackage{multibib}
\usepackage{svg}
\usepackage[utf8]{inputenc} % allow utf-8 input
\usepackage[T1]{fontenc}    % use 8-bit T1 fonts
\usepackage{hyperref}       % hyperlinks
\usepackage{url}            % simple URL typesetting
\usepackage{booktabs}       % professional-quality tables
\usepackage{amsfonts}       % blackboard math symbols
\usepackage{nicefrac}       % compact symbols for 1/2, etc.
\usepackage{microtype}      % microtypography
\usepackage{tikz}
\usetikzlibrary{arrows}
\usetikzlibrary{backgrounds}
\usepackage{graphicx}
\renewcommand{\beq}{\begin{equation}}
\renewcommand{\eeq}{\end{equation}}

\renewcommand{\eqref}[1]{equation(\ref{#1})}

\usepackage{mdframed}
\usepackage{caption}

\title{Integrated Weak Learning}

\author{%
Peter Hayes$^{1,2}$ \quad Mingtian Zhang$^{1}$ \quad Raza Habib$^{1, 2}$ \quad Jordan Burgess$^{2}$ \\ \textbf{Emine Yilmaz}$^{1, 2}$ \quad \textbf{David Barber}$^{1, 2}$ \vspace{3mm}\\
$^1$Centre for Artificial Intelligence, University College London \quad $^2$ Humanloop\\
\texttt{\{p.hayes, m.zhang, d.barber, emine.yilmaz\}@cs.ucl.ac.uk}\\
\texttt{\{raza, jordan\}@humanloop.com}\\
}

\begin{document}

\maketitle

\begin{abstract}  

We introduce \textit{Integrated Weak Learning}, a principled framework that integrates weak supervision into the training process of machine learning models. Our approach jointly trains the end-model and a \textit{label model} that aggregates multiple sources of weak supervision. We introduce a label model that can learn to aggregate weak supervision sources differently for different datapoints and takes into consideration the performance of the end-model during training. We show that our approach outperforms existing weak learning techniques across a set of 6 benchmark classification datasets. When both a small amount of labeled data and weak supervision are present the increase in performance is both consistent and large, reliably getting a 2-5 point test F1 score gain over non-integrated methods.
% and as much as 20 point gain
% Code to re-produce all our experiments is available @  \url{https://anonymous.4open.science/r/iweld-74C7/}

% Weak Learning (WL) has emerged as a promising approach to reduce the amount of manual labeling required to train machine learning models. WL techniques generally consist of a \textit{label model} that aggregates multiple weak supervision sources and an \textit{end-model} that leverages these aggregated labels. 
% We introduce \textit{Integrated Weak Learning} (iWL) - a simple and principled framework that integrates the label model into the training procedure of the end-model. 
% Both models are trained jointly using Maximum Likelihood Learning.
% Our approach can learn to aggregate weak supervision sources differently for different data-points and takes into consideration performance of the end-model in this process. 
% Our approach out-performs existing WL approaches across a range of benchmark datasets, including the scenario where some strongly labeled data is made available.
\end{abstract}

% \vspace{-10pt}
\section{Introduction}

To overcome the cost of manual data annotation, it's become increasingly common to include cheaper but less reliable sources of supervision when training deep learning models \cite{wrench, ratner2017data, ratner2019training,snorke18drybell, ratner20triplet}. These noisy sources of supervision might include crowd labels, weaker models, distant supervision by knowledge bases, or manually curated heuristic rules, etc. \cite{ratner2017data}. Given relatively little reliably-annotated data and a set of weaker sources of noisy labels, how should one best combine them to train a supervised machine learning model?

Early attempts at answering this question \cite{ratner2017data} typically decompose the problem into two stages. They first consider how to form an estimate of the unobserved label given a set of noisy labels and then consider how to train an end-model on the denoised labels. This approach has had considerable practical success \cite{snorke18drybell, dunnmon19health}, enabling deep learning systems to be deployed in industry without manual labeling. Two-stage methods have the advantage that after denoising, the rest of the training pipeline remains essentially unchanged. To achieve this though, they make quite an unnatural independence assumption; they typically ignore the dependence of the approximate labels on the input data. The cost of this assumption is that valuable information from the end-model can not be incorporated in denoising and the estimated accuracy of different supervision sources is fixed across the entire dataset. In addition, in general, a small amount of task-specific \emph{manual} labels continues to be critical for reliable results \cite{snorke18drybell, goh2018using}, even when given access to pre-trained models \cite{sun17revisit}. 

In this paper we present a probabilistic modeling framework, \emph{Integrated Weak Learning} (iWL), that simultaneously denoises the weak supervision sources and trains the end-model. Though there have been other approaches to joint model training and denoising \cite{salva21weasel, ren20denoise}, they have typically relied on intuitive heuristics, required extensive changes to the training objective and have high performance variance in our experiments.

Our approach is based on a simple application of maximum likelihood learning in an appropriate graphical model. It can learn to exploit the relevant expertise of different weak supervision sources, whilst taking into consideration the performance of the end-model. It consistently outperforms existing state-of-the-art weak learning techniques, both one-stage and two-stage, across 6 benchmark classification datasets of varying size and complexity.

\section{Weak Supervision}

In traditional supervised learning we have access to a set of labeled data pairs $\{(x_1,y_1),\cdots, (x_N,y_N)\}$ that are identically and independently (i.i.d) sampled from a true underlying joint distribution $(x_n,y_n)\sim p_d(x,y)=p_d(y|x)p(x)$. We are interested in applications where $y_{n}$ is expensive (in time and/or cost) to generate. A model $p_\theta(y|x)$ (for example a deep neural network with input $x$, output $y$ and parameters $\theta$) is then specified to approximate the true conditional distribution $p_d(y|x)$. In the classification setting, each label $y$ takes a discrete value in $\{1,\ldots, C\}$ where $C$ is the number of output classes for the problem. The parameter $\theta$ can be estimated by maximum likelihood estimation using the objective
\begin{equation}
    \theta^*=\arg\max_\theta \frac{1}{N} \sum_{n=1}^N \log p_\theta(y_n|x_n).
\end{equation}

In weak learning (WL), instead of requiring access to labeled data pairs, we assume access to $K$ weak supervision sources $\{\wl^1(\tilde{y}|x),\ldots,\wl^K(\tilde{y}|x)\}$. Each $\wl^{k}(\tilde{y}|x)$ can provide an approximate label $\tilde{y}^k$ given an input data $x$, resulting in a weakly labeled dataset $\sett{W}=\cb{(x_1,\tilde{y}^1_1,\ldots, \tilde{y}^K_1),\ldots,(x_W,\tilde{y}_W^1,\ldots,\tilde{y}_W^K)}$, where $W$ is the number of data-points that have weak labels only. In the data programming formalism \cite{ratner2017data} these supervision sources are encapsulated into \emph{labeling functions}. They usually can easily be applied across large unlabeled datasets. In principle, the domain of $\tilde{y}$ can be different from the domain of the true label $y$. For example, in our experiments we use labeling functions that can return one of $C+1$ classes $\tilde{y}\in\{0,\ldots, C\}$, where the class 0 represents an abstain where the labeling function refrains from making a decision.

% Intuitively, if the weak labels $\tilde{y}$ are reasonably approximations to the true label $y$, then the weak labels $\tilde{y}$ may serve as a useful surrogate for learning the model parametrs,$\theta$. 

If the weak labels have moderate accuracy then they can still serve as a useful source of supervision for training an end-model. Each of the labeling functions may not be very informative about the true label across all data points, but collectively these sources may help accurately predict the true label. The simplest approach to aggregating weak supervision sources is to take the majority vote label from the weak labels for each datapoint. Popular WL approaches improve on this by training a probabilistic model that learns an estimate for the accuracy of each labeling function and use this to calculate the distribution $p(y |\tilde{y}^1_{n},\ldots, \tilde{y}^K_{n})$ \cite{ratner17snorkel, ratner2019training}. 

In real world applications it is common for the practitioner to have access to both weak labels and a (relatively small) set of training data with `strong' labels (i.e. labels we assume come from the true underlying data distribution, such as those generated by human subject-matter experts). With this in mind, we denote the dataset that contains both strong and weak labels as  $\sett{S}=\cb{(x_1,y_1,\tilde{y}^1_1,\ldots, \tilde{y}^K_1),\ldots,(x_S,y_S,\tilde{y}^1_S,\ldots, \tilde{y}^K_S)}$, where $y_s\sim p_d(y|x_s)$. 

In the next section we present a simple probabilistic graphical model that makes it possible to learn the parameters of the model $p_\theta(y|x)$ both when the only labels are from weak supervision and when there are some strong labels present.

\section{Integrating Weak Supervision into Model Training}
\label{sec:iwl}

%  Given the presence of a correct underlying label $y$ for input $x$, with a set of weak labels $\tilde{y}$, we posit that a natural model for the weak labels $\wl(\tilde{y}|x)$ is

Given a weakly supervised dataset $W$, we wish to specify a probabilistic model over the weak label $\tilde{y}$ and the input $x$, that would allow us to jointly learn an end-model $p_\theta(y|x)$ and denoise the weak labels. A simple approach would be to parameterize the full distribution over an observed weak label $\tilde{y}$ and an input $x$ without loss of generality as: 

\beq
p_{\theta,\phi}(\tilde{y}|x) = \sum_y p_\phi(\tilde{y}|y, x)p_\theta(y|x),
\eeq

where we have marginalized over the unobserved true label $y$. However, this simple model has a potentially degenerate solution in which the distribution $p_\phi(\tilde{y}|y, x)$ becomes independent or approximately independent of the label $y$, i.e $p_\phi(\tilde{y}|y, x) \approx p_\phi(\tilde{y}|x)$. If that were to happen, then there would be no sharing of information between the weak labels and the true label $y$. In order to ensure that the weak labels contribute to the training of the end-model, we must constrain the distribution $p_\phi(\tilde{y}|y, x)$ so that the information flow from $x$ to $\tilde{y}$ is limited. The simplest way to achieve this constraint is to introduce an assumption that $\tilde{y}$ is independent of $x$ given $y$:

\beq
p_{\theta,\phi}(\tilde{y}|x) = \sum_y p_\phi(\tilde{y}|y)p_\theta(y|x).
\eeq

We refer to $p_\phi(\tilde{y}|y)$ as the label model and $p_\theta(y|x)$ is free to be any parametric end-model.

\begin{figure}[h]
     \centering
     \begin{subfigure}[b]{0.40\textwidth}
         \centering
        \begin{tikzpicture}[latent/.append style={minimum size=0.8cm},obs/.append style={minimum size=0.8cm}]
         \node[obs] (x) {$x_n$}; %
         \node[latent,below=of x] (y) {$y_n$}; %
         \node[obs,below=of y] (y_tilde) {$\tilde{y}^k_{n}$}; %
          \node[latent, right=of y_tilde] (phi) {$\phi^k$};
          \node[latent, left=of x] (theta) {$\theta$};
        \plate {plate1} {(y_tilde)} {$K$};
        \plate {plate2} {(phi)} {$K$};
        \plate {} {(x)(y)(plate1)} {$N$};
        % edges
         \edge {x} {y}
         \edge {y} {y_tilde}
         \edge {theta} {y}
         \edge {phi} {y_tilde}
         \end{tikzpicture}
         \caption{Basic graphical model (iWL)}
         \label{fig:graphical_model_basic}
     \end{subfigure}
     \begin{subfigure}[b]{0.40\textwidth}
         \centering
         \begin{tikzpicture}[latent/.append style={minimum size=0.8cm},obs/.append style={minimum size=0.8cm}]
         \node[obs] (x) {$x_n$}; %
         \node[latent,below=of x] (y) {$y_n$}; %
         \node[obs,below=of y] (y_tilde) {$\tilde{y}^k_{n}$}; %
          \node[latent, right=of y_tilde] (phi) {$\phi^k_n$};
          \node[latent, left=of x] (theta) {$\theta$};
        %\node[latent, above=of phi] (eta) {$\eta$};
        \plate {plate1} {(y_tilde)(phi)} {$K$};
        \plate {} {(x)(y)(phi)(plate1)} {$N$};
        % edges
         \edge {x} {y}
         \edge {y} {y_tilde}
         \edge {x}{phi}
         \edge {theta} {y}
         \edge {phi} {y_tilde}
         %\edge {eta} {phi}
         \end{tikzpicture}
         \caption{Including $x$ dependency (iWLD)}
         \label{fig:graphical_model_adv}
     \end{subfigure}
        \caption{Graphical models for the integrated weak learning variants. $\theta$ and $\Phi=\{\phi^1,\ldots,\phi^K\}$ are the parameters of the end-model and label model respectively. $K$ is the number of available weak supervision sources and $N$ the number of observed training datapoints. The left  model assumes the generative process is the same for all datapoints, whereas the right model assumes similar datapoints will have a similar noisy label generative process. }
        \label{fig:graphical_model}
\end{figure}
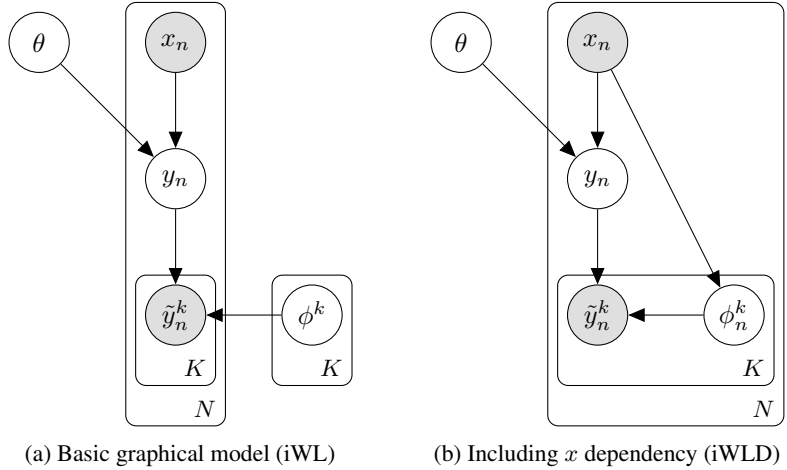

This model formulation allows us to jointly estimate the parameters $\{\Phi,\theta\}$ using maximum likelihood estimation. In practice we have $K$ weak supervision sources. If we assume these sources are conditionally independent given $y$, our model becomes
\beq
p_{\theta,\Phi}(\tilde{y}^1,\ldots,\tilde{y}^K|x) = \prod_{k=1}^K\sum_y p_{\phi^k}(\tilde{y}^k|y)p_\theta(y|x),\label{eq:simple_label_model}
\eeq
where we denote $\Phi=\{\phi^1,\ldots,\phi^K\}$.
Figure \ref{fig:graphical_model_basic} shows the corresponding graphical model. 

The log-likelihood for the data-points that include both strong and weak labels $\sett{S}=\cb{(x_1,y_1,\tilde{y}^1_1,\ldots, \tilde{y}^K_1),\ldots,(x_S,y_S,\tilde{y}^1_S,\ldots, \tilde{y}^K_S)}$ becomes:
\beq
L^S(\theta,\Phi) \equiv \frac{1}{S}\sum_{s=1}^S \br{\log p_\theta(y_s|x_s) + \sum_{k=1}^K\log \sum_{y_s} p_{\phi^k}(\tilde{y}^k_s|y_s)p_\theta(y_s|x_s)} 
\label{eq:lik:strong}.
\eeq
For the data-points that only have weak labels $\sett{W}=\cb{(x_1,\tilde{y}^1_1,\ldots, \tilde{y}^K_1),\ldots,(x_W,\tilde{y}_W^1,\ldots,\tilde{y}_W^K)}$ we have the log-likelihood term
\beq
L^W(\theta,\Phi) \equiv \frac{1}{W}\sum_{w=1}^W  \sum_{k=1}^K\log \sum_{y_w} p_{\phi^k}(\tilde{y}^k_w|y_w)p_\theta(y_w|x_w)
\label{eq:lik:weal}.
\eeq
Therefore, our overall training objective for integrating weak supervision (iWL) into model training is
\beq
L(\theta,\Phi)=\lambda_S L^S(\theta,\Phi)+\lambda_W L^W(\theta,\Phi),
\eeq
\label{eq:objective}
where $\lambda_S,\lambda_L$ are user defined scaling parameters that control how much the strongly and weakly labeled datasets influence the model training. By jointly estimating both sets of parameters $\{\Phi,\theta\}$ using this objective, the label model can learn to take into consideration the performance of the end-model during training. In practice we use stochastic gradient descent to train the parameters, which requires different mini-batches of data of the same size to be sampled for the $L^{S}$ and $L^{W}$ terms for each parameter update - see section \ref{sec:experiments} for further details. Once trained, the discriminator $p_\theta(y|x)$ can be used independently to the label model to make predictions. This framework is flexible for supporting different choices of label and end-model.

% \david{I don't think its good to call the parameterisations "linear". This is a bit confusing and potentially misleading. They are just distributions and not functions which are linear in their parameters. I would strongly suggest to remove the word "linear" throughout.}
% \david{N.B "in practice", not "in practise"}

\subsection{Design of Label Model}
\label{sec:label_model}

In the previous section, we assumed that the weak labels only depend on the underlying true label $y$, yielding the label model in equation \ref{eq:objective}. In this case $p_{\phi^k}(\tilde{y}|y)$ is parameterized by a linear transition matrix i.e.  $p_{\phi^k}(\tilde{y}^k=i|y=j)=\phi^k_{ij}$, 
with $i\in\{1,\ldots,C+1\}$, including the abstain label, and $j\in\{1,\ldots,C\}$. Each column in the transition matrix sums to one: $\sum_{i}\phi_{ij}^{k}=1$. 

% In principle, we could also explore a nonlinear transition and leverage Expectation Maximization (EM) to learn the parameters, which we leave for future work - see appendix \ref{app:EM} for more information.

A natural extension is to consider incorporating a dependency on $x$. The label model could then represent different transitions for different data-points. Intuitively this would allow the label model to understand the relevant expertise of the different labeling functions and emphasize appropriately. This is particularly relevant given in practice labeling functions tend to be quite specialized in the data-points they perform well on \cite{wrench}. However, we must take great care when introducing $x$-dependence to constrain the flow of information so that the label model does not become independent of $y$.

We therefore propose to parameterize the transition matrix $\phi^{k}$ itself as a function of $x$ - see figure \ref{fig:graphical_model_adv} for the updated graphical model. This ensures that $p(\tilde{y}|x, y)$ is still parameterized by a linear transition matrix.  The label model becomes $p(\tilde{y}^{k}|y,\phi^{k}=f_\eta(x)[k])$, where
$f_\eta(\cdot)$ is a neural network that maps a data point $x$ to the $K$ linear transition matrices and we use $f_\eta(x)[k]$ to denote the $k$th matrix. This allows our label model to produce similar transitions $\phi^{k}$ for similar $x$. Therefore, the full model can be written as 
\begin{align}
    p_{\eta,\theta}(\tilde{y}^1,\ldots \tilde{y}^K|x)=  \prod_{k=1}^K \sum_y p(\tilde{y}^k|y, \phi^k=f_\eta(x)[k]) p_\theta(y|x). \label{eq:amortized:transition}
\end{align}

In principle, it's still possible though unlikely that the label model learns to ignore $y$. In section \ref{sec:experiments} we validate that this does not happen in practice and demonstrate how incorporating the $x$ dependency in this way improves performance across a range of benchmark problems.

As with existing weak learning techniques, it is possible to extend our label model to more explicitly consider correlations between labeling functions - see appendix \ref{app:correlations} for more details.

\subsection{Analysis of Model}
\label{sec:analysis}
As with prior works \cite{ratner17snorkel}, our model can suffer from $y$ being unidentifiable. Fortunately, in our framework, access to strongly labeled data naturally mitigates against this. For notational simplicity we consider the case of only one labeling function $\wl(\tilde{y}|x)=\sum_y p_{\phi^*}(\tilde{y}|y)p_{d}(y|x)$ where $p_d(y|x)$ is the true label generation distribution and we assume $p_{\phi^*}(\tilde{y}|y)$ parameterized by a linear transition matrix, i.e. $p_{\phi^*}(\tilde{y}=i|y=j)=\phi^*_{ij}$ with known parameters $\phi^*$.
In this case, training our model $p_\theta(y|x)$ using equation \ref{eq:objective} is able to identify the true underlying $p_d(y|x)$. Specifically, for each data-point $x$, maximizing the likelihood function is equivalent to minimizing the KL divergence 
    \beq
       \mathbb{E}_{\wl(\tilde{y}|x)}(\log p_{\theta,\phi^*}(\tilde{y}|x)) =-\mathrm{KL}\left(\sum_y p_{\phi^*}(\tilde{y}|y)p_{d}(y|x)|| \sum_y p_{\phi^*}(\tilde{y}|y)p_{\theta}(y|x)\right)+const.
    \eeq
During training, as $\theta\rightarrow\theta^*$ the KL divergence goes to 0 and we have  
  \beq
      \sum_j\phi^*_{ij}p_{d}(y=j|x)=\sum_j\phi^*_{ij}p_{\theta^*}(y=j|x) \quad \forall i.
  \eeq
As long as the linear transition matrix $\phi^*$ (with size $C+1\times C$) does not degenerate (i.e. has rank $C$), then the mapping from the distribution of $y$ to $\tilde{y}$ is injective. We then have $p_{\theta^*}(y|x)=p_{d}(y|x)$, which successfully recovers the underlying true model. 
This result  generalizes the identifiablity theorem of the discrete spread divergence\footnote{In the discrete spread divergence~\cite{zhang2020spread}, the considered transition matrix is an invertible square matrix, whereas the transition matrix in this work is a rectangle and assumed to be an injective mapping.} (see Appendix B of \cite{zhang2020spread}). To note, when $\phi$ is a function of $x$ (\ref{fig:graphical_model_adv}) the same reasoning can be applied if we  assume the true transition generation function $f_{\eta^*}$ is also known.

When the true label model $p_{\phi^*}(\tilde{y}|y)$ is unknown, we need to learn $\phi$. In this case we cannot guarantee to identify the true underlying model using only the weak labels. To give an example, we first assume that we have learned a label model $p_{\phi_1}(\tilde{y}|y)$ using MLE with a model $p_{\theta_1}(y|x)$. We can always construct an alternative label model $p_{\phi_2}(\tilde{y}|y)$ (with $\phi_2=\phi_1\times m^{-1}$, where $m$ is a $C\times C$ invertible transition matrix, and model $p_{\theta_2}(y|x)$ (with $p_{\theta_2}(y=i|x)=\sum _j m_{ij} p_{\theta_1}(y=j|x)$) that can give the same marginal distribution:
\begin{align}
    \sum_y p_{\phi_1}(\tilde{y}|y)p_{\theta_1}(y|x)=\sum_y p_{\phi_2}(\tilde{y}|y)p_{\theta_2}(y|x).
\end{align}
This observation easily generalizes to the case of $K$ weak supervision sources. Similar counter examples are constructed in the context of disentangled representation learning, where the true representation cannot be identified under maximum likelihood learning~\cite{locatello2019challenging}.

Incorporating strong labels $y\sim p_{d}(y|x)$, like we discuss in section \ref{sec:iwl}, can alleviate this unidentifiability issue. Intuitively, when the number of strong labels goes to infinity $S\rightarrow\infty$, then the first term in 
Equation \ref{eq:lik:strong} $\frac{1}{S}\sum_{s=1}^S \log p_\theta(y_s|x_s)$ will reach an optimum when $p_\theta(y|x)=p_d(y|x)$, which is due to the consistency of MLE~\cite{casella2021statistical}. In practice, we find that a relatively small number of strong labels is effective at mitigating this issue allowing us to recover a useful end-model. We leave to future work exploring more deeply the relationship between the identifiability problem and the number of strong labels required. In the case where it is not possible to access any strong labels for a given problem, similar to prior work we can leverage the majority vote heuristic to initialize the parameters of our label model, which we find works well empirically. 

\section{Related work}
\label{sec:related}

\textbf{Two-Stage Weak Learning:} Two stage weak labeling methods separate the label model from end-model training \cite{ratner2017data, ratner17snorkel, ratner2019training, ratner20triplet} . The primary advantage of this separation is that the cost of label denoising is paid only once and the change needed to training pipelines is minimal.

In our work it is necessary to alter the training objective by adding additional terms and one has to learn the parameters of the label model every time the end-model is changed. However, the additional computational cost of learning the label model $p_\phi(\tilde{y}|y)$ will usually be small compared to the cost of training most modern deep learning end-models $p_\theta(y|x)$. Our experiments demonstrate that this additional cost is consistently rewarded by superior end-models, especially when some strong labels are present.
% Most prior-work has focused on improving the label model. For example, learning the structure of labeling function dependencies \cite{ratner17structure}, extending to efficiently incorporate multi-task weak supervision\cite{ratner18multitask} and improving inference speed and scalability \cite{ratner20triplet}. 
% In the first stage, the label model is generally represented by a probabilistic graphical model that also treats $y$ as latent and learns an estimate for the accuracy of each labeling function and then uses this to calculate the distribution $p(y |\tilde{y}^1_{n},\ldots, \tilde{y}^K_{n})$ . Numerous techniques have been introduced to enhance this label model stage. . Once learned, the estimated label distribution is then generally used to separately train the end-model with a noise-aware likelihood objective. Unlike this family of approaches, our method jointly learns both the label model and end-model and so the label model can consider the performance of the end-model during training, improving overall performance. Furthermore, we propose a label model that can learn to consider the expertise of different labeling functions. 

\textbf{Joint Weak Learning:} Most similar in spirit to our work are two end-to-end weak labeling methods that also jointly denoise and train: \emph{WeaSel} \cite{salva21weasel} and \emph{Denoise} \cite{ren20denoise}. These methods differ both in how they parameterize the label model and in their training objective. The primary difference with our work is in the choice of training objective. Whereas we train using maximum-likelihood learning, Weasel uses a heuristic consistency constraint. Namely that the labels predicted by a denoising model and an end-model should agree. Training their heuristic objective can be unstable \cite{salva21weasel} and can result in degenerate solutions. In their paper the method is primarily justified by empirical performance but in our experiments it under-performed both Denoise and Integrated Weak Learning (see section \ref{sec:experiments}). In contrast, our framework simply proposes an appropriate graphical model and trains via maximum likelihood with stochastic gradient descent.

% They differ from our work in their choice of label model and overall learning objective. The label models of both Denoise and WeaSEL are neural network models that take as input the data and weak labels $(x_{n},\tilde{y}^1_{n},\ldots, \tilde{y}^K_{n})$ and output an accuracy (also referred to as reliability) weighting $A_{n} = [a_{n}^{1}, \ldots, a_{n}^{K}]$ over the $K$ weak labels for each datapoint, where $\sum_{k}a_n^{k} = 1 $. Estimates for the true label are then calculated simply using an accuracy weighted vote of the weak labels for each datapoint. These estimates are then used as observed targets for the end-model in their co-training style objective. Denoise recommends to use an attention based neural network, whereas WeaSEL promote their approach as being model agnostic.

% proposes a probabilistic graphical model that encodes our belief about how the data was generated, assuming latent $y$ and modeling the weak label $\tilde{y}$ as a stochastic generation given the true label $y$. Unlike these other end-to-end methods, this allows us to integrate weak supervision into a principled maximum likelihood objective that uses only the observed strong and weak labels, which is consistent and asymptotically efficient~\cite{casella2021statistical}. In addition, in section \ref{sec:experiments}, we empirically show that our proposed framework outperforms both Denoise and WeaSEL in the majority of cases. 

The Denoise algorithm has an additional algorithmic component beyond weak supervision which incorporates self-supervision \cite{yarowsky95selfsuper, karamanolakis21selfsup} of its end-model. Confident predictions from the end-model are bootstrapped for learning where the labeling functions have low coverage. Based on their ablation studies, this significantly improves performance. Self-supervision could naturally be applied to our objective \ref{eq:objective} to further bolster performance, which we leave for future work.

\textbf{Learning with Noisy Labels:}
There are multiple different approaches to learning with noisy labels, including data cleaning \cite{lang20noisycleaning, zhang20noisyclean, yang2018distantly}, where useful information is potentially lost, and data re-weighting that weighs training data-points based based on different criteria \cite{zhang21incompletesup, ren18reweightdl}. 
Most relevant to our line of work are those methods that attempt to correct noisy labels using a label model parameterized similarly to ours. In \emph{Confident Learning} \cite{northcutt2021confident} they assume a probabilistic label model, treating the true underlying $y$ as a latent, and do inference on a single $C \times C$ noise transition matrix in order to correct their noisy labels. Our approach differs in a few key ways. Firstly, our label model is integrated as part of the end-model training. Secondly, we are dealing with multiple transition matrices, one per each source of label noise, where the noisy label domain is different than the true label domain. Lastly, our label model from figure \ref{fig:graphical_model_adv} can learn different transitions for different $x_{n}$, releasing the assumption that label noise needs to be constant across a dataset. In \cite{wang2019learning} they do in fact have a noise transition matrix that is dependent on $x_{n}$, but they have an alternating training scheme for the end-model, instead of jointly training, and they too don't deal with multiple sources of noisy labels with different domains.

\section{Experiments}
\label{sec:experiments}

The goal of our experiments is to provide a robust performance comparison between the variants of our iWL approach and the existing weak learning approaches discussed in section \ref{sec:related}. In addition, we want to understand how the amount of strongly labeled data $\mathcal{L}$ impacts the performance of these methods for deep learning models (which may be of independent interest for practitioners). For our choice of datasets and the implementations of existing methods, we leverage the recent comprehensive benchmark for weak supervision (WRENCH) \cite{wrench}. Specifically, we use 6 of the classification problems that vary in dataset size as well as labeling function complexity - see table \ref{tab:datasets} for details. We compare to the two-stage weak learning approaches of Majority Vote and Snorkel \cite{ratner2017data}, and to the end-to-end weak learning approaches of Denoise \cite{ren20denoise} and the more recent WeaSEL \cite{salva21weasel}. We refer to these as benchmark methods. We use the implementations of these methods available in the WRENCH benchmark. 
% Code to reproduce all our experiments is available:  \url{https://anonymous.4open.science/r/iweld-74C7/}.

\begin{table}[ht]
    \caption{Attributes of the chosen classification datsets \cite{wrench}}
    \small
    \centering
    \setlength\tabcolsep{4pt}
    \begin{tabular}{r c  c  c  c c }
    \toprule
        \textbf{Dataset} &  \textbf{\#Classes} & \textbf{\#LFs} & \textbf{\#Train} & \textbf{\#Validation} & \textbf{\#Test}\\
        \hline
                Census   &  2  & 83 & 10,083 & 5,561 & 16,281\\
                IMDB     & 2  & 5 & 20,000 & 2,500 & 16,281\\
                Yelp     & 2  & 8 & 30,400 & 3,800 & 3,800\\
                SMS      & 2  & 73 & 4,751 & 500 & 500\\
                AGNews   & 4  & 9 & 96,000 & 12,000 & 12000\\
                TREC     & 6  & 68 & 4,965 & 500 & 500\\
        \bottomrule
        \vspace{2pt}
    \end{tabular}
    \label{tab:datasets}
\end{table}

\subsection{Implementation Details}
\label{sec:experiments:detail}
Here we introduce our basic implementation details. Further details can be found in appendix \ref{app:experiment_details}.
    
\textbf{Discriminative model:} We keep the end discriminative model $p_{\theta}(y|x)$ common across all methods in our comparisons and vary the label model accordingly. Specifically for $p_{\theta}(y|x)$ we use the distilled RoBerta transformer model \cite{sanh2019distilbert} to provide a rich embedded representation for textual $x$. This acts as input to a two-layer feed forward neural network model, with 100 hidden units in each layer and RelU activation functions and a softmax final output. Across all experiments we use the Adam optimizer \cite{kingma2014adam} with learning rate $1e^{-3}$ and mini-batch size of 128. As in WRENCH we do early stopping on the validation F1-score with a patience of 300 optimization iterations using the validation datasets provided. Specific to our proposed integrated approach: we set the hyperparameters $\lambda_{L}, \lambda_{W}$ from equation \ref{eq:objective} to 1 throughout - meaning we weight equally the contributions from the strong and weak labels in our objective. For the two-stage weak learning approaches of majority vote and snorkel, we use the probabilistic denoised labels (as opposed to one-hot) and noise-aware loss objective as recommended in prior works.

\textbf{Label model:} 
For our proposed approach we include both of the label model variants presented in figure \ref{fig:graphical_model} - including and excluding the dependency on $x$. We refer to these as iWL and iWLD respectively in the results. For iWLD, the network $f_{\eta}(x): x \rightarrow \{\phi^{1},\ldots, \phi^{K}\}$ uses the same architecture and hyperparameters as $p_{\theta}(y|x)$ as specified above, except for the structure of the final layer that instead outputs the linear transition matrix. We initialize our label model parameters using the majority vote to mitigate against the non-identifiability of $y$ issue discussed in section \ref{sec:analysis}. For the benchmark methods we use the default label model hyperparameters as provided by WRENCH except for the WeaSEL model. For the WeaSEL temperature hyperparameter we try values from the range $\{0.5, 1, 3, 5\}$ based on their recommendations and select the best performing value for each experiment configuration because we found this method to be sensitive to this parameter in our setup. In appendix \ref{app:viz_transitions} we provide visualizations to illustrate how our label model is able to learn different transitions $\phi_{n}^{k}$ for different datapoints.

\textbf{Incorporating strong labels:} We also evaluate all methods with different proportions of strong labels available at training time in addition to the weak labels. We believe this to be a realistic and important scenario for many real world applications. Specifically we evaluate the scenarios where $1\%$, $10\%$, $50\%$ and $100\%$ of the training data is strongly labeled (selected at random). The strong label log-likelihood term in our objective equation \ref{eq:lik:strong} means that our proposed approach can deal with this scenario by design. To ensure that the benchmark methods also benefit from these strong labels in our comparisons, we add an additional labeling function in these cases that outputs the strong label if available and abstains otherwise. For the two-stage weak learning approaches, we also include results for an alternative approach of leaving the labeling functions unchanged and instead replacing the resulting denoised label with the corresponding strong label if available when training the end-model $p_{\theta}(y|x)$  - see appendix~\ref{app:ablations} for further details.

\begin{figure}[t]
    \centering
    \begin{subfigure}[b]{0.45\textwidth}
        \centering
        \includegraphics[width=1.1\linewidth]{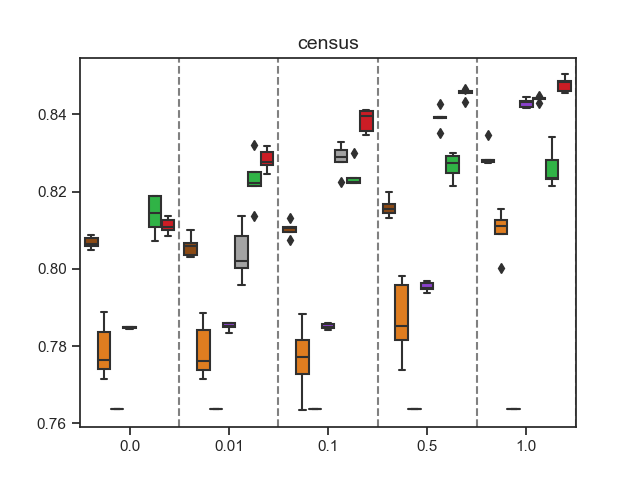}
    \end{subfigure}%
    \begin{subfigure}[b]{0.45\textwidth}
        \centering
        \includegraphics[width=1.1\linewidth]{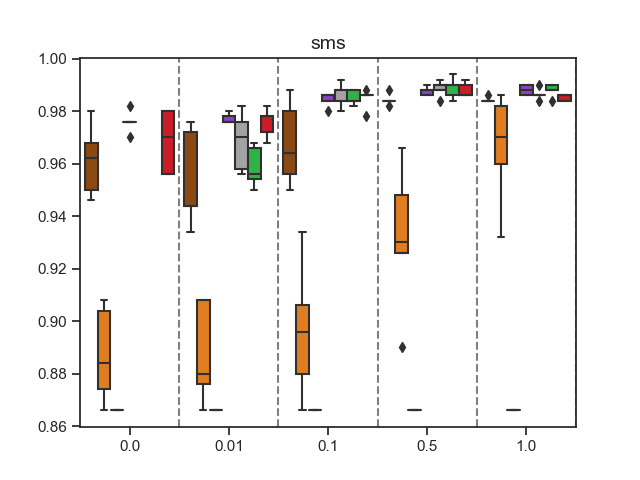}
    \end{subfigure} \\
    \vspace{-10pt}
    \begin{subfigure}[b]{0.45\textwidth}
        \centering
        \includegraphics[width=1.1\linewidth]{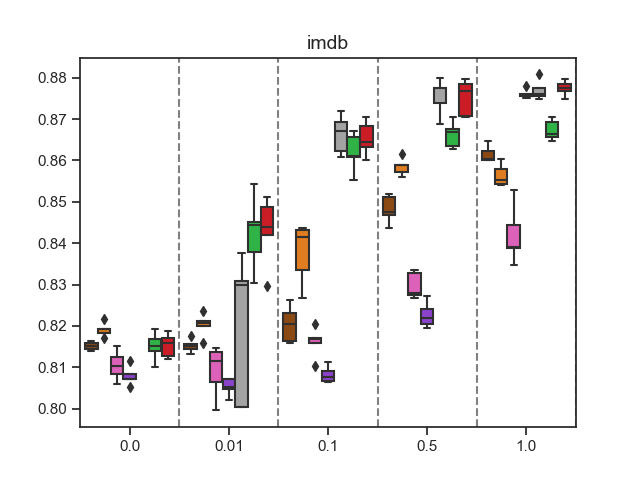}
    \end{subfigure}
    \begin{subfigure}[b]{0.45\textwidth}
        \centering
        \includegraphics[width=1.1\linewidth]{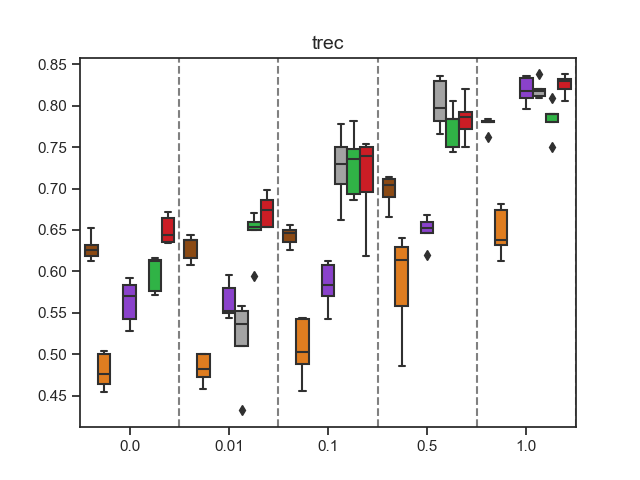}
    \end{subfigure} \\
    \vspace{-10pt}
    \begin{subfigure}[b]{0.45\textwidth}
        \centering
        \includegraphics[width=1.1\linewidth]{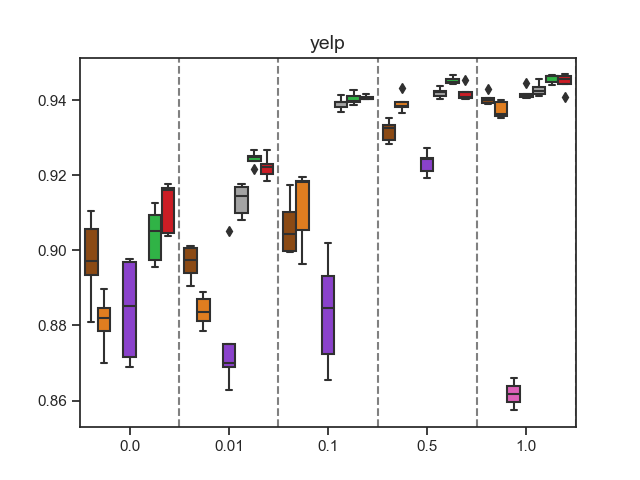}
    \end{subfigure}%
    \begin{subfigure}[b]{0.45\textwidth}
        \centering
        \includegraphics[width=1.1\linewidth]{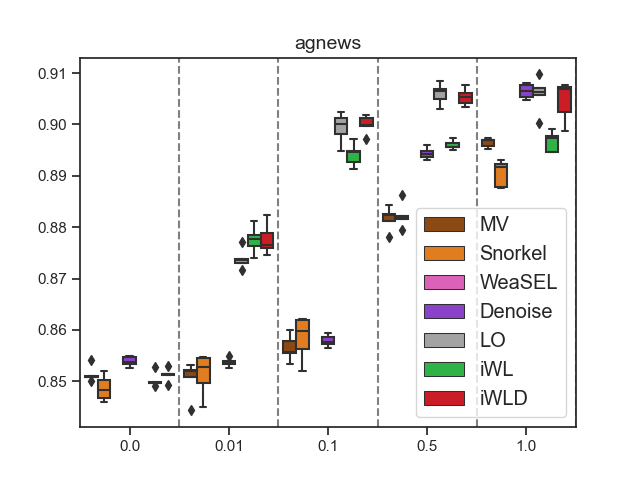}
    \end{subfigure}
    \caption{Box-plots of test F1 scores showing the quantiles across 5 random seeds on the y-axis. Each plot refers to a specific dataset. The models are grouped by the different fractions of strongly labeled data available on the x-axis ($0\%$, $1\%$, $10\%$, $50\%$, $100\%$). In the majority of datasets and strong label splits, our proposed iWL and iWLD models are the best performing models and are robust to random seed changes shown by the relatively low variance. Methods that converge to significantly worse or degenerate solutions fall below a performance display threshold and are not shown. See table \ref{tab:main_results} for more detail.}
    \label{fig:fig-boxplots}
\end{figure}
\subsection{Results}

We report the test F1-score as our main evaluation metric averaged over 5 random seeds alongside 1 standard deviation. In table \ref{tab:main_results} we summarize the performance results for the different methods, for different proportions of strongly labeled data, across all datasets considered. In figure \ref{fig:fig-boxplots}, we plot the corresponding quartiles of the F1 scores to provide further insight into how the variability between seeds compares across the different methods. We also include a baseline where we train the end-model $p_{\theta}(y|x)$ only on the strong labels provided, ignoring any weak labels. We refer to this baseline as `labels only' (LO).

We see from table \ref{tab:main_results} that our proposed approaches result in the best performing model (as measured by test F1 score averaged over 5 random seeds) in 19 out of the 30 cases and first or second best-performing in 28 out of the 30 cases. In particular, in the regime where you have a relatively small amount of strongly labeled data in addition to your weak labeling sources (i.e. the $1\%$ row in table \ref{tab:main_results}) our approach provides a conclusive improvement in all but one of the datasets. Here we consistently outperform the weak learning baselines by between 2 and 5 test-F1 points. In 5 of the 6 datasets we see that our integrated weak learning approach outperforms the LO baseline of the end-model trained with 100\% strong labels available. Furthermore, the boxplot quartiles in figure \ref{fig:fig-boxplots} illustrates that for a majority of the datsets and strong label splits, our iWLD model results in lower variance solutions than the benchmark methods.  

Our results also provide some noteworthy insights related to the benchmark methods. Generating results across a range of different strong label proportions demonstrates that there is a tipping point at $10\%$ over which LO becomes a competitive baseline. This can still represent a relatively significant amount of labeling effort (e.g. in AGNews this would be $9600$ labels). It is likely that our use of the distilled RoBerta transformer as the feature extractor will be contributing to this performance, bringing some transfer learning benefits. 

The joint approach Denoise is highlighted as the best performing approach in a small number of cases. This was unexpected because in the original WRENCH benchmark Denoise failed to outperform the other methods in any of these datasets.  Furthermore, the more recent WeaSEL paper does not compare to Denoise as an end-to-end alternative. We note that Denoise, in addition to the weak and strong labels provided, also incorporates self-supervision signal into their training process. Our framework can naturally be extended with self-supervised labels which will likely further improve performance. Finally, we experienced that WeaSEL failed with degenerate solutions in some of the experiments, for example for the TREC dataset, and performed for the most part worse than other methods. In an attempt to improve performance for WeaSEL, we tuned the temperature parameter as discussed in section \ref{sec:experiments:detail}.

\begin{table}[ht]
\caption{Test F1 score averaged over 5 random seeds with 1 standard deviation in brackets across all datasets and all models considered: two stage weak learning approaches of Majority Vote (MV) and Snorkel. End-to-end weak learning approaches of WeaSEL and Denoise. The end-model $p_\theta(y|x)$ trained with the available strong labels only (labels only - LO). Our proposed integrated training approach with and without the $x$ dependency in the label model (iWL and iWLD respectively). The results are grouped by the different proportions of strongly labeled data available ($0\%$, $1\%$, $10\%$, $50\%$, $100\%$). Values highlighted in \textbf{\red{red}} indicates best performing and \textbf{\blue{blue}} indicates second best. Our proposed approaches are the best performing models as measured by average test F1 score in 19 out of the 30 cases and the first or second best performing models in 28 out of the 30 cases.}
    \centering
    \small
    \setlength\tabcolsep{4pt}
    \setlength\extrarowheight{3pt} 
          \begin{tabular}{llllllll}
    \toprule
         & \textbf{Dataset} &        AGNews &        Census &          IMDB &           SMS &          TREC &          Yelp \\
    \textbf{Labels} & \textbf{Model} &               &               &               &               &               &               \\
    \toprule
    \multirow{7}{*}{0\%} & MV &  \blue{\textbf{85.14} (0.16)} &  80.68 (0.16) &  81.42 (0.10) &  96.12 (1.38) &  62.80 (1.54) &  89.75 (1.16) \\
         & Snorkel &  83.15 (3.85) &  77.89 (0.72) &  80.91 (0.16) &  88.72 (1.84) &  47.96 (2.19) &  88.09 (0.74) \\
         & WeaSEL &  66.32 (1.64) &  76.38 (0.00) &  81.05 (0.35) &  86.60 (0.00) &  27.60 (0.00) &  54.38 (2.04) \\
         & Denoise &  \red{\textbf{85.39} (0.10)} &  78.48 (0.02) &  80.79 (0.24) &  \red{\textbf{97.60} (0.42)} &  56.32 (2.74) &  88.41 (1.35) \\
         & LO &    - &    - &    - &    - &    -  &    - \\
         \cline{2-8}
         & iWL &  85.02 (0.14) &  \red{\textbf{81.40} (0.51)}  & \blue{\textbf{81.50} (0.34)} &  62.32 (2.04) &  \blue{\textbf{59.80}(2.20)} &  \blue{\textbf{90.41} (0.74)} \\
         & iWLD &  85.13 (0.13) &  \blue{\textbf{81.11} (0.20)} &  \red{\textbf{81.54} (0.29)} &  \blue{\textbf{96.84} (1.20)} &  \red{\textbf{65.00} (1.71)} &   \red{\textbf{91.17} (0.69)} \\
    \midrule
    \multirow{7}{*}{1\%} & MV &  85.05 (0.36) &  80.59 (0.28) &  81.52 (0.16) &  95.40 (1.88) &  62.88 (1.58) &  89.67 (0.45) \\
         & Snorkel &  83.45 (3.78) &  77.89 (0.73) &  82.03 (0.28) &  88.76 (1.93) &  48.24 (1.82) &  88.38 (0.43) \\
         & WeaSEL &  66.30 (2.10) &  76.38 (0.00) &  80.92 (0.63) &  86.60 (0.00) &  27.60 (0.00) &  53.62 (1.34) \\
         & Denoise &  85.37 (0.09) &  78.51 (0.11) &  80.53 (0.21) &  \red{\textbf{97.72} (0.18)} &  56.44 (2.25) &  87.64 (1.67) \\
         & LO &  87.39 (0.20) &  80.41 (0.71) &  81.98 (1.80) &  96.84 (1.13) &  51.76 (5.13) &  91.34 (0.42) \\
         \cline{2-8}
         & iWL &  \blue{\textbf{87.75} (0.27)} &  \blue{\textbf{82.29} (0.67)} &  \blue{\textbf{84.25} (0.89)} &  95.88 (0.78) &  \blue{\textbf{64.56} (2.98)} &  \red{\textbf{92.43} (0.18)} \\
         & iWLD &  \red{\textbf{87.77} (0.31)} &  \red{\textbf{82.82} (0.29)} &  \red{\textbf{84.31} (0.84)}&  \blue{\textbf{97.44} (0.55)} &  \red{\textbf{67.32} (1.95)} &  \blue{\textbf{92.21} (0.31)} \\
    \midrule
    \multirow{7}{*}{10\%} & MV &  85.65 (0.25) &  81.03 (0.21) &  82.05 (0.44) &  96.76 (1.60) &  64.28 (1.19) &  90.63 (0.76) \\
         & Snorkel &  84.80 (2.36) &  77.67 (0.94) &  83.78 (0.73) &  89.64 (2.60) &  50.64 (3.73) &  91.16 (1.03) \\
         & WeaSEL &  68.42 (0.49) &  76.38 (0.00) &  81.62 (0.36) &  86.60 (0.00) &  27.60 (0.00) &  61.02 (8.33) \\
         & Denoise &  85.79 (0.12) &  78.51 (0.07) &  80.82 (0.20) &  98.44 (0.26) &  58.32 (2.88) &  88.35 (1.48) \\
         & LO &  \blue{\textbf{89.94} (0.30)} &  \blue{\textbf{82.86} (0.39)} &  \red{\textbf{86.63} (0.47)} &  \red{\textbf{98.64} (0.46)}&  \blue{\textbf{72.52} (4.41)} &  93.91 (0.17) \\
         \cline{2-8}
         & iWL &  89.42 (0.22) &  82.42 (0.33) &  86.20 (0.47) &  98.52 (0.27) &  \red{\textbf{72.92} (3.97)}&  \blue{\textbf{94.03} (0.16)} \\
         & iWLD &  \red{\textbf{90.00} (0.18)} &  \red{\textbf{83.84} (0.30)} &  \blue{\textbf{86.53} (0.42)} &  \blue{\textbf{98.48} (0.39)} &  71.16 (5.72) &  \red{\textbf{94.06} (0.06)} \\
    \midrule
    \multirow{7}{*}{50\%} & MV &  88.17 (0.22) &  81.59 (0.26) &  84.82 (0.34) &  98.44 (0.22) &  69.72 (1.98) &  93.18 (0.28) \\
         & Snorkel &  88.23 (0.24) &  78.69 (1.01) &  85.85 (0.21) &  93.20 (2.84) &  58.56 (6.41) &  93.92 (0.25) \\
         & WeaSEL &  70.47 (2.17) &  76.38 (0.00) &  82.98 (0.32) &  86.60 (0.00) &  27.60 (0.00) &  71.89 (4.14) \\
         & Denoise &  89.44 (0.11) &  79.53 (0.13) &  82.26 (0.31) &  98.76 (0.17) &  64.92 (1.83) &  92.33 (0.32) \\
         & LO &  \red{\textbf{90.60}} (0.21) &  \blue{\textbf{83.91} (0.27)} &  \red{\textbf{87.56}} (0.44) &  \red{\textbf{98.88} (0.30)} &  \red{\textbf{80.24} (3.02)} &  \blue{\textbf{94.19} (0.13)} \\
         \cline{2-8}
         & iWL &  89.60 (0.09) &  82.66 (0.34) &  86.62 (0.31) &  \red{\textbf{98.88} (0.39)} &  77.36 (2.60) &  \red{\textbf{94.51} (0.10)} \\
         & iWLD &  \blue{\textbf{90.53} (0.17)} &  \red{\textbf{84.55} (0.13)} &  \blue{\textbf{87.52} (0.43)} &  \blue{\textbf{98.80} (0.28)} & \blue{\textbf{78.40}(2.58)} &  94.18 (0.21) \\
    \midrule
    \multirow{7}{*}{100\%} & MV &  89.65 (0.09) &  82.93 (0.30) &  86.15 (0.21) &  98.44 (0.09) &  77.76 (0.89) &  94.03 (0.16) \\
         & Snorkel &  89.05 (0.25) &  80.97 (0.58) &  85.64 (0.27) &  96.60 (2.16) &  64.76 (2.95) &  93.74 (0.22) \\
         & WeaSEL &  72.72 (3.04) &  76.38 (0.00) &  84.20 (0.69) &  86.60 (0.00) &  27.60 (0.00) &  84.43 (1.81) \\
         & Denoise & \red{\textbf{90.65} (0.14)} &  84.29 (0.11) &  87.61 (0.11) &  \red{\textbf{98.80} (0.23)} &  81.88 (1.68) &  94.17 (0.16) \\
         & LO &  \blue{\textbf{90.59} (0.34)} &  \blue{\textbf{84.40} (0.07)} &  \blue{\textbf{87.70} (0.24)} &  \blue{\textbf{98.64} (0.22)} &  \blue{\textbf{81.96} (1.11)} &  94.28 (0.18) \\
         \cline{2-8}
         & iWL &  89.67 (0.20) &  82.61 (0.51) &  86.73 (0.24) &  \red{\textbf{98.80} (0.24)} &  78.40 (2.19) &  \red{\textbf{94.56} (0.12)} \\
         & iWLD &  90.46 (0.39) &  \red{\textbf{84.78} (0.20)} &  \red{\textbf{87.74} (0.18)} &  98.52 (0.11) &  \red{\textbf{82.52} (1.25)} &  \blue{\textbf{94.47} (0.24)} \\
    \bottomrule
    \end{tabular}
\end{table}
\label{tab:main_results}

\section{Conclusion}

We have proposed a new framework for training supervised machine learning models that can principally integrate both strong and weak supervision sources during training. It models the true underlying label $y$ as a latent variable and jointly trains both the label model and end-model parameters using maximum likelihood. It is a generic framework that can be used in conjunction with existing supervised learning models to improve performance. We provide an extensive empirical study across a range of classification benchmark problems of varying degrees of size and complexity and demonstrate that our approach consistently outperforms existing state-of-the-art methods.

% This work acts as a foundation for future work in many complimentary directions. Firstly, improving the label model to explicitly model correlations between weak supervision sources. We outline a provisional approach based on introducing an additional latent variable in appendix \ref{app:correlations}. Secondly, exploring alternative parameterizations for our label model as presented in \ref{sec:label_model}, including those that introduce constraints that remove the need to initialize with majority vote in the cases where no strong labels are available. Lastly, understanding how to combine our framework with related methods that attempt to mitigate the burden of gathering labeled data - namely self-supervised learning, active learning and transfer learning.

% \begin{ack}

% \end{ack}

\clearpage
\bibliography{main.bib}
\bibliographystyle{abbrvnat}

\newpage

\appendix

\section{Further ablation}
\label{app:ablations}
\begin{figure}[ht]
    \centering
    \begin{subfigure}[b]{0.45\textwidth}
        \centering
        \includegraphics[width=1.1\linewidth]{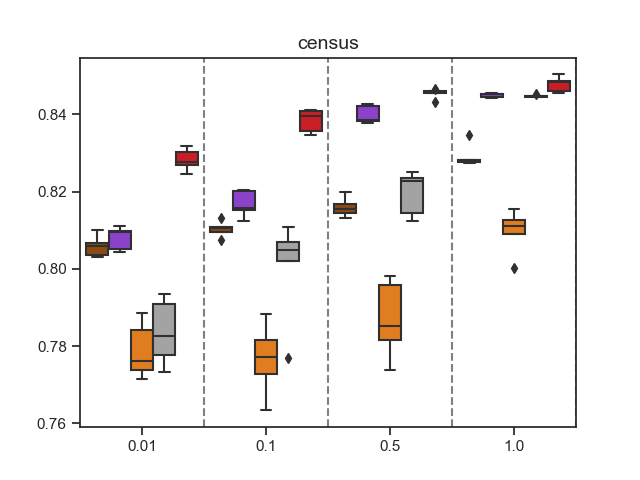}
    \end{subfigure}%
    \begin{subfigure}[b]{0.45\textwidth}
        \centering
        \includegraphics[width=1.1\linewidth]{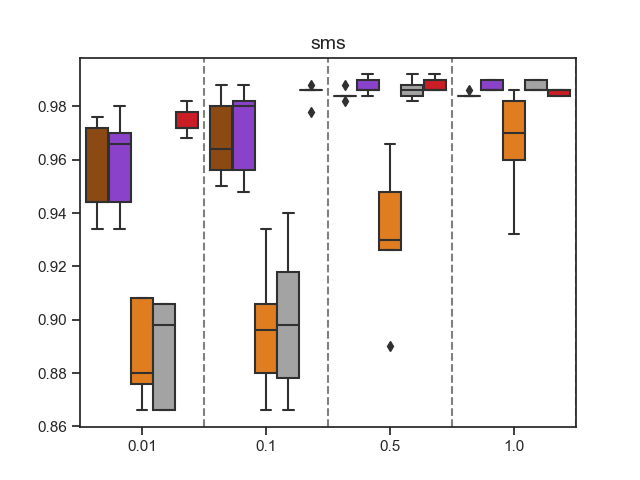}
    \end{subfigure} \\
    \vspace{-10pt}
    \begin{subfigure}[b]{0.45\textwidth}
        \centering
        \includegraphics[width=1.1\linewidth]{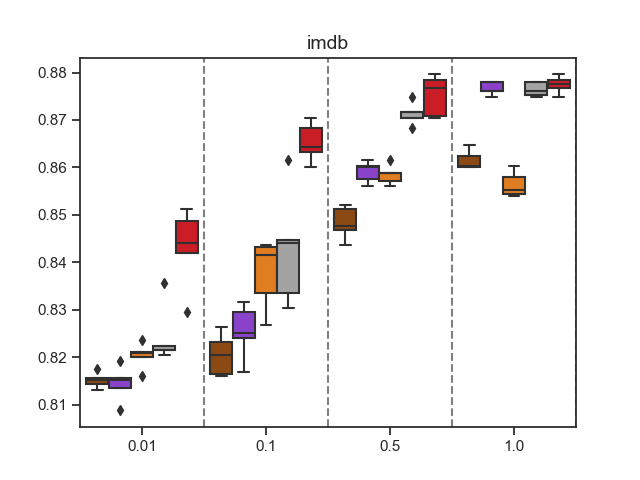}
    \end{subfigure}
    \begin{subfigure}[b]{0.45\textwidth}
        \centering
        \includegraphics[width=1.1\linewidth]{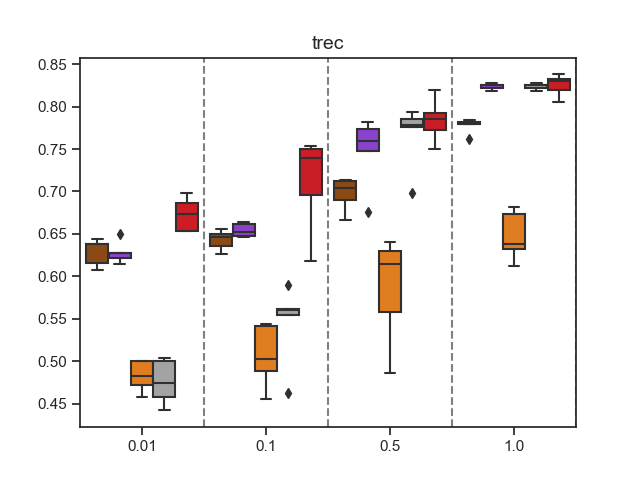}
    \end{subfigure} \\
    \vspace{-10pt}
    \begin{subfigure}[b]{0.45\textwidth}
        \centering
        \includegraphics[width=1.1\linewidth]{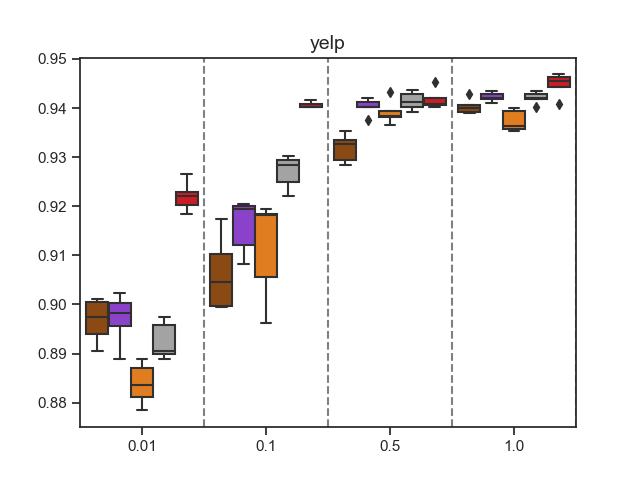}
    \end{subfigure}%
    \begin{subfigure}[b]{0.45\textwidth}
        \centering
        \includegraphics[width=1.1\linewidth]{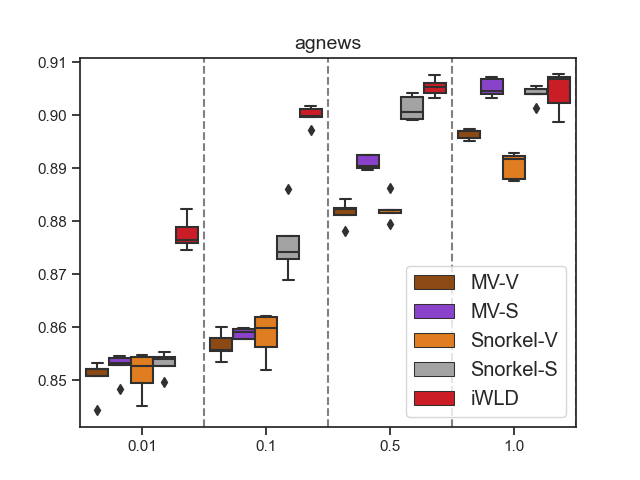}
    \end{subfigure}
    \caption{Box-plots of test F1 scores showing the quantiles across 5 random seeds on the y-axis. Each plot refers to a specific dataset. The model variants discussed in section \ref{app:ablations} are grouped by the different fractions of strongly labeled data available on the x-axis ($1\%$, $10\%$, $50\%$, $100\%$). See table \ref{tab:ablation} for more detail.}
    \label{fig:fig-boxplots-ablation}
\end{figure}

In this section we provide results for the ablation on how to include strong labels when available in the two-stage baseline methods, namely Snorkel and Majority Vote (MV). The standard approach is to add an additional labeling function that outputs the strong label if available and abstains otherwise. We refer to this as the voting variant (-V). This is what we use in the results reported in section \ref{sec:experiments}. An alternative approach we also consider is to leave the labeling functions unchanged and instead replace the resulting denoised label with the corresponding strong label if available when training the end-model $p_{\theta}(y|x)$. We refer to this as the strong variant (-S). Figure \ref{fig:fig-boxplots-ablation} and the corresponding table \ref{tab:ablation} compare Snorkel-V, Snorkel-S, MV-V and MV-S to our model variant iWLD across the same 6 datasets for different splits of available strongly labeled data. We can see that the -S variants in fact consistently outperform the -V variants. Furthermore our iWLD approach is the best performing method in 28 out of the 30 cases.

\begin{table}[ht]
\caption{Test F1 score averaged over 5 random seeds with 1 standard deviation in brackets across all datasets and model variants discussed in section \ref{app:ablations}. The results are grouped by the different proportions of strongly labeled data available ($1\%$, $10\%$, $50\%$, $100\%$). Values highlighted with \textbf{bold} indicate the best performing method.}
    \centering
    \small
    \setlength\tabcolsep{4pt}
    \begin{tabular}{rlllllll}
    \toprule
         &  &        Agnews &        Census &          IMDB &           SMS &          TREC &          Yelp \\
    Labels & Model &               &               &               &               &               &               \\
    \midrule
    \multirow{5}{*}{1\%} & MV-V &  85.05 (0.36) &  80.59 (0.28) &  81.52 (0.16) &  95.40 (1.88) &  62.88 (1.58) &  89.67 (0.45) \\
         & MV-S &  85.27 (0.25) &  80.80 (0.30) &  81.45 (0.38) &  95.88 (1.91) &  62.84 (1.34) &  89.71 (0.52) \\
         & SKL-V &  83.45 (3.78) &  77.89 (0.73) &  82.03 (0.28) &  88.76 (1.93) &  48.24 (1.82) &  88.38 (0.43) \\
         & SKL-S &  84.23 (2.47) &  78.36 (0.85) &  82.43 (0.63) &  88.84 (2.07) &  47.56 (2.67) &  89.25 (0.38) \\
         & iWLD &  \textbf{87.77} (0.31) &  \textbf{82.82} (0.29) &  \textbf{84.31} (0.84) &  \textbf{97.44} (0.55) &  \textbf{67.32} (1.95) &  \textbf{92.21} (0.31) \\
    \midrule
    \multirow{5}{*}{10\%} & MV-V &  85.65 (0.25) &  81.03 (0.21) &  82.05 (0.44) &  96.76 (1.60) &  64.28 (1.19) &  90.63 (0.76) \\
         & MV-S &  85.89 (0.10) &  81.68 (0.34) &  82.54 (0.57) &  97.08 (1.76) &  65.44 (0.82) &  91.61 (0.56) \\
         & SKL-V &  84.80 (2.36) &  77.67 (0.94) &  83.78 (0.73) &  89.64 (2.60) &  50.64 (3.73) &  91.16 (1.03) \\
         & SKL-S &  87.59 (0.64) &  80.03 (1.34) &  84.29 (1.22) &  90.00 (2.99) &  54.56 (4.88) &  92.71 (0.34) \\
         & iWLD &  \textbf{90.00} (0.18) &  \textbf{83.84} (0.30) &  \textbf{86.53} (0.42) &  \textbf{98.48} (0.39) &  \textbf{71.16} (5.72) &  \textbf{94.06} (0.06) \\
    \midrule
    \multirow{5}{*}{50\%} & MV-V &  88.17 (0.22) &  81.59 (0.26) &  84.82 (0.34) &  98.44 (0.22) &  69.72 (1.98) &  93.18 (0.28) \\
         & MV-S &  89.11 (0.14) &  83.99 (0.23) &  85.91 (0.23) &  98.76 (0.33) &  74.80 (4.23) &  94.03 (0.17) \\
         & SKL-V &  88.23 (0.24) &  78.69 (1.01) &  85.85 (0.21) &  93.20 (2.84) &  58.56 (6.41) &  93.92 (0.25) \\
         & SKL-S &  90.14 (0.24) &  81.96 (0.57) &  87.14 (0.23) &  98.64 (0.38) &  76.64 (3.89) &  94.15 (0.18) \\
         & iWLD &  \textbf{90.53} (0.17) &  \textbf{84.55} (0.13) &  \textbf{87.52} (0.43) &  \textbf{98.80} (0.28) &  \textbf{78.40} (2.58) &  \textbf{94.18} (0.21) \\
    \midrule
    \multirow{5}{*}{100\%} & MV-V &  89.65 (0.09) &  82.93 (0.30) &  86.15 (0.21) &  98.44 (0.09) &  77.76 (0.89) &  94.03 (0.16) \\
         & MV-S &  \textbf{90.53} (0.18) &  84.50 (0.06) &  87.66 (0.14) &  \textbf{98.76} (0.22) &  82.40 (0.40) &  94.23 (0.09) \\
         & SKL-V &  89.05 (0.25) &  80.97 (0.58) &  85.64 (0.27) &  96.60 (2.16) &  64.76 (2.95) &  93.74 (0.22) \\
         & SKL-S &  90.40 (0.16) &  84.48 (0.04) &  87.64 (0.15) &  \textbf{98.76} (0.22) &  82.40 (0.40) &  94.21 (0.12) \\
         & iWLD &  90.46 (0.39) &  \textbf{84.78} (0.20) &  \textbf{87.74} (0.18) &  98.52 (0.11) &  \textbf{82.52} (1.25) &  \textbf{94.47} (0.24) \\
    \bottomrule
    \end{tabular}
\end{table}
\label{tab:ablation}

\section{Further experiment details}
\label{app:experiment_details}

All experiments were run on a GPU cluster with access to 10 V100 GPU processors. Each experiment run was executed on a single GPU instance. The code we have made available for producing our experimental results is implemented using PyTorch \cite{pytorch}. 

The specific train, validation and test data splits used across all datasets are available in a standardized schema at the following URL: \url{https://drive.google.com/drive/folders/1VFJeVCvckD5-qAd5Sdln4k4zJoryiEun}. This is provided as part of the WRENCH benchmark \cite{wrench}. Further information on how this data was gathered, the original source, and the relevant attributes is available in WRENCH \cite{wrench}.

\section{Visualizing the label model}
\label{app:viz_transitions}
\begin{figure}[ht]
    \centering
    \begin{subfigure}[b]{0.33\textwidth}
        \centering
        \includegraphics[width=1.3\linewidth]{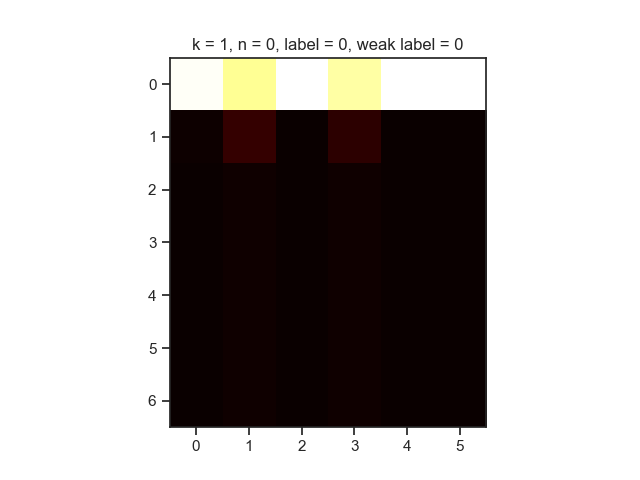}
    \end{subfigure}%
    \begin{subfigure}[b]{0.33\textwidth}
        \centering
        \includegraphics[width=1.3\linewidth]{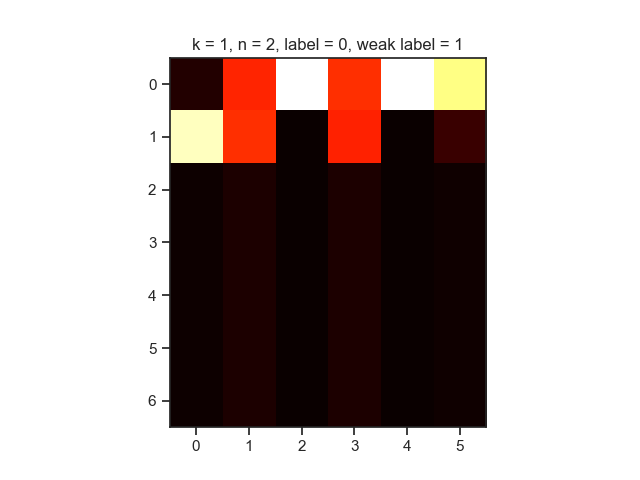}
    \end{subfigure} 
    \begin{subfigure}[b]{0.33\textwidth}
        \centering
        \includegraphics[width=1.3\linewidth]{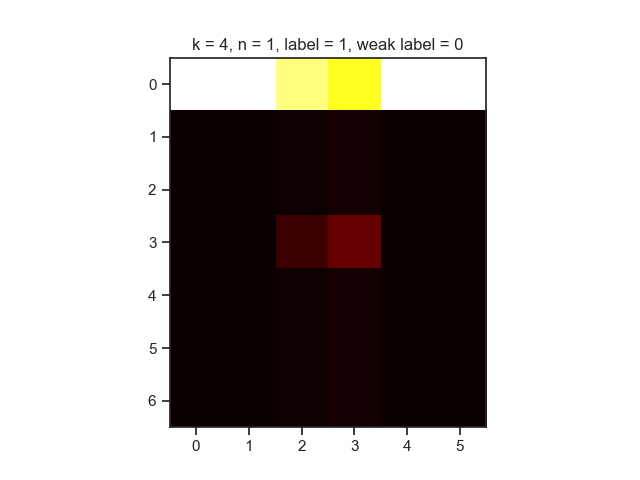}
    \end{subfigure} \\
        \begin{subfigure}[b]{0.33\textwidth}
        \centering
        \includegraphics[width=1.3\linewidth]{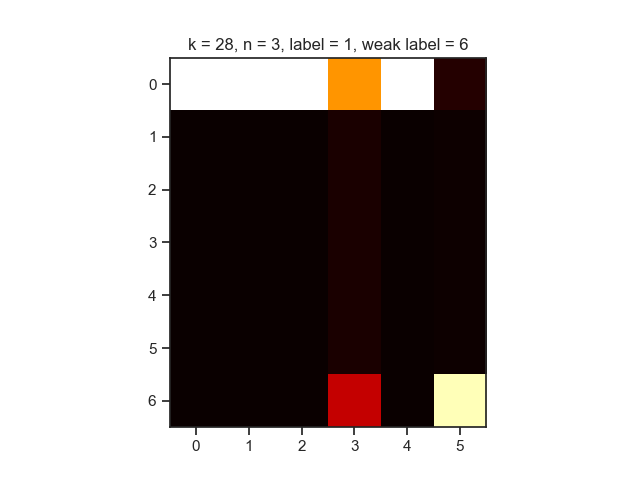}
    \end{subfigure}%
    \begin{subfigure}[b]{0.33\textwidth}
        \centering
        \includegraphics[width=1.3\linewidth]{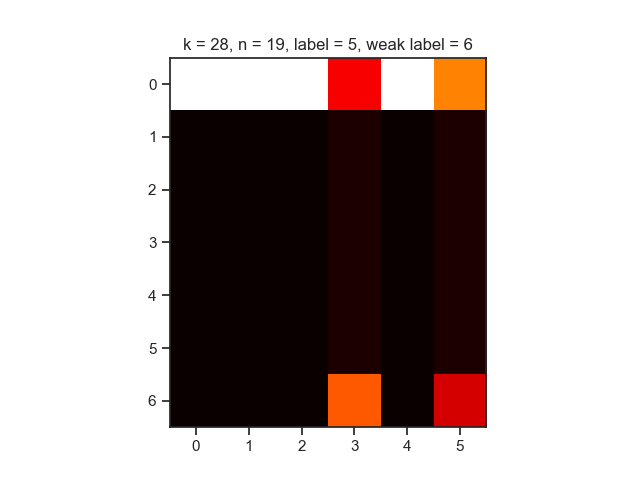}
    \end{subfigure} 
    \begin{subfigure}[b]{0.33\textwidth}
        \centering
        \includegraphics[width=1.3\linewidth]{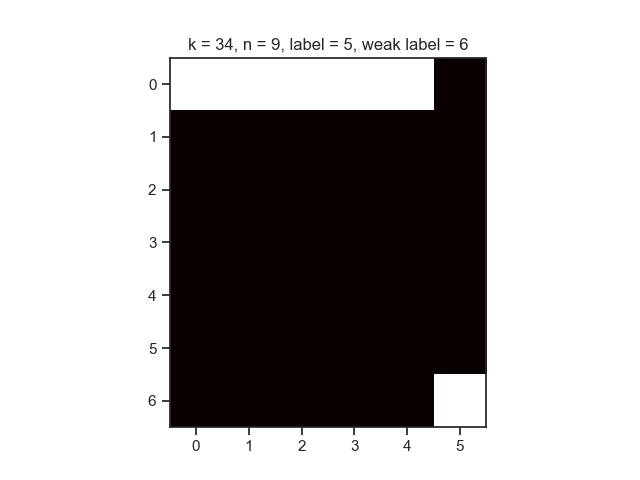}
    \end{subfigure} \\
           \begin{subfigure}[b]{0.33\textwidth}
        \centering
        \includegraphics[width=1.3\linewidth]{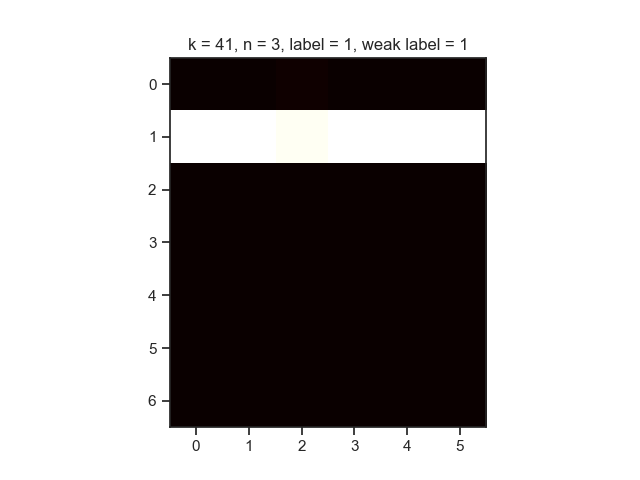}
    \end{subfigure}%
    \begin{subfigure}[b]{0.33\textwidth}
        \centering
        \includegraphics[width=1.3\linewidth]{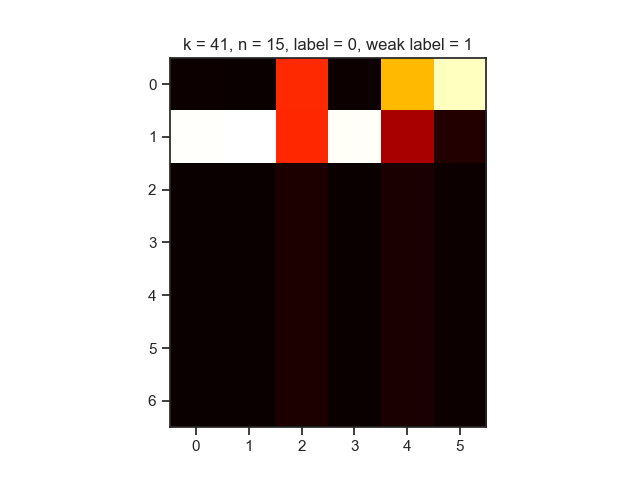}
    \end{subfigure} 
    \begin{subfigure}[b]{0.33\textwidth}
        \centering
        \includegraphics[width=1.3\linewidth]{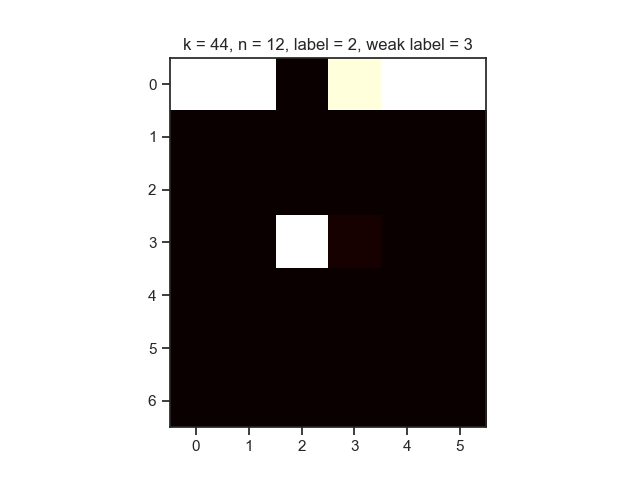}
    \end{subfigure} \\
    \caption{Heat plots of the transitions $\phi_{n}^{k}$ at convergence after training. For an array of different labeling functions $k$ and different datapoints $n$, sampled from the TREC dataset (number of class labels $C=6$ and number of labeling functions $K=68$). This illustrates that our model is learning different transitions for different datapoints.}
    \label{fig:visualization}
\end{figure}

In section \ref{sec:label_model} we proposed a label model that included a dependency on $x$ (see figure \ref{fig:graphical_model_adv}). Figure \ref{fig:visualization} plots an array of different transitions $\phi_{n}^{k}$, after training, for different labeling functions $k$ and different datapoints $n$. These are sampled from the TREC dataset (number of class labels $C=6$ and number of labeling functions $K=68$). This helps illustrate that our model has in fact learned different transitions for different datapoints.

\section{Dependent labeling functions}
\label{app:correlations}

There is a straightforward extension to our model that may be well suited to where we have multiple dependent labels $\tilde{y}_1,\ldots,\tilde{y}_K$. In principle, the labeling functions are conditionally independent given $x$. To take information from the weak labels, our assumption in section \ref{sec:iwl} is that we can explain the weak labels based only on the true label $y$, that is 
\beq
p_\phi(\tilde{y}|y,x) = \prod_{k=1}^K p_\phi(\tilde{y}_k|y).
\eeq
A simple alternative choice to consider dependent weak labels is to include an additional latent $h$
\beq
p_\phi(\tilde{y}|y,x) = \sum_h p_\phi(h|x)\prod_{k=1}^K p_\phi(\tilde{y}_k|y,h).
\eeq
For a discrete $h$ we can use the EM algorithm for learning. In principle, we can also include a dependency $p_\phi(h|y)$. We leave further investigation of this variant to future work.

\end{document}